\documentclass[runningheads]{llncs}
\usepackage[T1]{fontenc}

\usepackage[english]{babel}
\usepackage[letterpaper,top=2cm,bottom=2cm,left=3cm,right=3cm,marginparwidth=1.75cm]{geometry}
\usepackage{booktabs}
\usepackage{subcaption}

\usepackage{amsmath,amsfonts}
\usepackage{graphicx}
\usepackage[colorlinks=true, allcolors=blue]{hyperref}
\usepackage{cleveref}

\title{PrivacyGuard: A Modular Framework for Privacy Auditing in Machine Learning}

\author{
Luca Melis\inst{1} \and 
Matthew Grange\inst{1} \and 
Iden Kalemaj\inst{1} \and
Karan Chadha\inst{1} \and
Shengyuan Hu\inst{1} \and
Elena Kashtelyan\inst{1} \and
Will Bullock\inst{1}
}
\authorrunning{L. Melis et al.}
\institute{Meta.}

\begin{document}
\maketitle

\begin{abstract}
The increasing deployment of Machine Learning (ML) models in sensitive domains motivates the need for robust, practical privacy assessment tools.
PrivacyGuard is a comprehensive tool for empirical differential privacy (DP) analysis, designed to evaluate privacy risks in ML models through state-of-the-art inference attacks and advanced privacy measurement techniques. To this end, PrivacyGuard implements a diverse suite of privacy attack-- including membership inference , extraction, and reconstruction attacks -- enabling both off-the-shelf and highly configurable privacy analyses. Its modular architecture allows for the seamless integration of new attacks, and privacy metrics, supporting rapid adaptation to emerging research advances. We make PrivacyGuard available at \url{https://github.com/facebookresearch/PrivacyGuard}.
\end{abstract}

\section{Background and Introduction}
\label{sec: intro}
The widespread deployment of Machine Learning (ML) models across sensitive applications has increased the concerns about the privacy risks associated with these models. 
Modern ML systems, particularly those operating on large-scale training data pipelines with neural networks, can inadvertly memorize and expose sensitive information from their training dataset.
The urgency of addressing these privacy risks has been amplified by evolving regulatory landscapes and AI governance frameworks. Modern data protection legislation, notably the European Union's General Data Protection Regulation (GDPR)~\cite{gdpr2016} establishes stringent requirements for organizations deploying AI systems that process personal information. Specifically, GPDR article 35 requires data protection impact assessments that quantitatively analyze and minimize privacy risks throughout AI development lifecycles.

Traditional approaches to privacy protection, most notably Differential Privacy (DP), provide strong theoretical guarantees by injecting carefully calibrated noise into the training algorithm~\cite{dwork2014algorithmic}.
However, these worst-case privacy bounds often prove to be overly conservative in practice, significantly limiting model utility, while potentially failing to capture the nuanced privacy risks of real world deployments~\cite{jayaraman2019evaluating}.

Empirical DP~\cite{burchard2019empirical} (eDP) has emerged as a practical \emph{privacy guardrail} that addresses these limitations by measuring the actual privacy loss through empirical evaluation rather than just a theoretical worst-case analysis as in DP.
Instead of relying on noise injection and mathematical bounds, eDP employs privacy attacks --- such as Memberships Inference Attacks~\cite{shokri2017membership} (MIAs) -- to directly asses whether an adversary can determine if specific data-points were included in a model's training set. This approach enables a lower bound on the privacy risk assessment while unlocking greater model utility. In recent years, this approach has been widely adopted in both academic research (cite) as well as industry applications~\cite{kong2024dp,agrawal2025quantifying}. 

To facilitate the adoption and further research of these techniques, we present \emph{PrivacyGuard}, a comprehensive PyTorch-based framework for conducting State-Of-The-Art (SOTA) empirical privacy analysis of ML models. PrivacyGuard addresses the critical gap between theoretical privacy research and practical privacy assessment by providing a unified, modular platform that implements a diverse set of privacy attacks methodologies and integrates seamlessly with training ML as well as privacy attacks workflows. Our frameworks supports both traditional supervised learning models and modern generative AI systems, including Large Language Models (LLMs), thus making it applicable across the full spectrum of modern ML applications.

As we discuss in~\Cref{sec:design}, the technical architecture of PrivacyGuard is built around a modular design pattern that separates the privacy attacks from privacy analyses, with the aim of enabling researchers and practitioners to easily configure, extend and customize their privacy evaluations. The framework implements a comprehensive suite of privacy attacks thorough specialized analysis nodes.
Each of these attacks is implemented as a configurable module that inherits from a base analysis interface, promoting code reuse and enabling rapid prototyping of new methodologies.

For Membership Inference Attacks, we provide both lightweight calibration based attacks~\cite{calibmia} suitable for fast deployments as well as sophisticated SOTA attacks such as the Likelihood Ratio Attack (LiRA)~\cite{lira} and Robust MIA (RMIA~\cite{rmia}) that employ multiple auxiliary (shadow) models for enhanced attack performance.
For Generative models, we implement reconstruction attacks that evaluate and adversary's ability to extract PII given access to masked training data, as well as pragmatic extraction methods that simulate real-world adversarial scenarios. 
We provide tutorials in the form of python notebooks on the CIFAR-10~\cite{cifar10} and Enron~\cite{klimt2004enron} datasets.

The framework computes key privacy metrics derived from the attack performance like extraction rates and Area Under the Curve (AUC) for average-case privacy assessment as well as empirical epsilon to capture worst-case privacy bounds at different attack error thresholds. These metrics provide a quantitative way of evaluating the balance between privacy protection with model utility. We discuss privacy analysis methodologies used in PrivacyGuard in~\Cref{sec:auditing_mia}. We provide tutorials on how to leverage our privacy analysis methodologies, including the recent work in~\cite{mahloujifar2025auditing} on efficiently auditing ML models in a single training run.

Finally in~\Cref{sec:related}, we discuss how PrivacyGuard compares with existing libraries in the context of privacy assessment of ML models. 




\section{Design Principles and Features}
\label{sec:design}
PrivacyGuard is built around a modular architecture that separates privacy attacks from privacy analyses, enabling researchers and practitioners to easily configure, extend, and customize their privacy evaluations. The framework's design follows two core principles: \textbf{modularity} through clear abstraction layers and \textbf{extensibility} to accommodate emerging privacy attack methodologies and analysis techniques.

\subsection{Privacy Attack Module}

\begin{figure}
    \centering
    \includegraphics[width=0.7\linewidth]{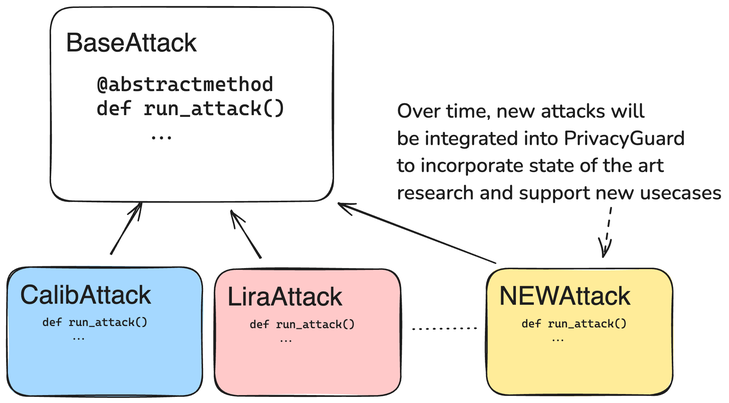}
    \caption{PrivacyGuard Attack module}
    \label{fig:design_base}
\end{figure}

The Privacy Attack Module serves as the central platform component for implementing a wide suite of privacy attacks across different use cases (\Cref{fig:design_base}). This generic module currently supports Membership Inference Attacks (MIA) and Label Inference Attacks, with an extensible architecture designed for continued addition of state-of-the-art privacy attacks, including emerging use cases in Generative AI.

The module provides three core functionalities:
\begin{itemize}
\item \textbf{Dataset Preparation:} The system can load and process datasets from multiple sources including files on disk, service APIs, and data pipelines, providing flexibility for different deployment scenarios.
\item \textbf{Attack Execution:} The module prepares response generation according to configurable sampling strategies defined by user parameters (e.g., temperature vs. greedy sampling, shadow model configuration, membership inference attack metrics). This allows researchers to experiment with different attack configurations while maintaining consistency across evaluations.
\item \textbf{Analysis Input Preparation:} Utilizing the results of executed attacks, the module prepares standardized ``AnalysisInput'' objects that are consumed by Analysis Nodes to provide meaningful insights into attack results. This standardization enables consistent analysis across all privacy attacks implemented in the framework.
\end{itemize}

All outputs follow a common format that allows for standardized analysis across different privacy attack methodologies. For example, in a typical Membership Inference Attack flow, the module takes assessment configuration and prompts/target verbatim as inputs, and produces token-level scores and generations as outputs, maintaining consistency across different attack implementations.
\subsection{Analysis Nodes}

\begin{figure}[ht]
    \centering
    \includegraphics[width=0.7\linewidth]{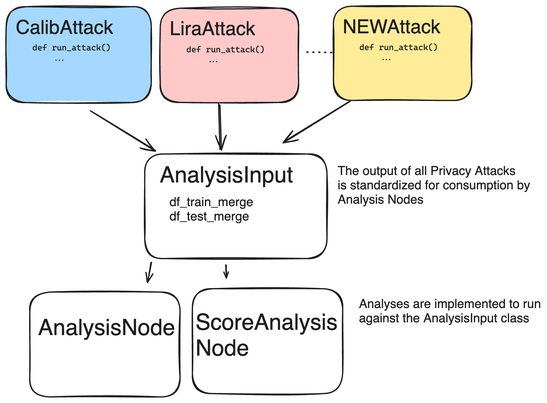}
    \caption{PrivacyGuard Analysis Node component}
    \label{fig:design_analysis}
\end{figure}
The Analysis Node component, illustrated in~\Cref{fig:design_analysis}, provides a generic interface for analyzing outputs from Privacy Attacks. Each AnalysisNode computes general metrics required to analyze the performance of a given model against executed attacks. The modular design allows users to select and configure the specific set of Analysis Nodes most relevant to their privacy attack use case.
\begin{itemize}
\item \textbf{Lightweight, Repeatable Process:} Using the ``AnalysisInput'' object produced by Privacy Attacks, the system can execute a variety of analyses through configurable analysis nodes. These are designed as lightweight, repeatable processes that can be executed with different configurations and thresholds, allowing for flexible deep dives into attack results without requiring re-execution of computationally expensive privacy attacks.
\item \textbf{Configurable Metric Analyses:} The framework provides a comprehensive suite of configurable metrics to evaluate attack results, including Area Under the Curve (AUC), empirical epsilon calculations, and extraction rates. This flexibility enables researchers to focus on the metrics most relevant to their specific privacy assessment goals.
\item \textbf{Report Generation:} Analysis Nodes prepare comprehensive metrics for automated report generation, including summary report cards that synthesize key findings across multiple attack methodologies.
\end{itemize}

\subsection{Robustness and Testing Framework}
PrivacyGuard is designed with robustness and correctness as primary considerations to ensure reliable performance in production code and workflows. The testing framework includes several key components:
\begin{itemize}
\item \textbf{Comprehensive Unit Testing:} The library maintains high code coverage (>90\% for core components) through extensive unit testing, ensuring individual components function correctly across different scenarios.
\item \textbf{End-to-End Validation:} The framework includes end-to-end testing and tutorials demonstrating privacy attacks executed against established datasets from academic literature, with analysis of their outputs to validate correctness against known baselines.
\item \textbf{Continuous Integration:} A robust CI/CD pipeline ensures that all library updates require successful test runs across the entire framework and its dependencies, maintaining code quality and preventing regressions.
\item \textbf{Type Safety:} Strict type checking using Pyre ensures code reliability and helps prevent runtime errors in production deployments.
\end{itemize}










\section{Auditing methodology and Results}
\label{sec:auditing_mia}

In this section, we provide an overview of the attacks and auditing analyses supported in PrivacyGuard. We provide examples of running these attacks on common datasets: CIFAR-10 for image classification and Enron for large language model fine-tuning. 

\subsection{Membership Inference Attacks on ML Models}

We demonstrate two SOTA attacks implemented in PrivacyGuard: Likelihood Ratio Attack (LiRA) \cite{lira} and  Robust MIA (RMIA) \cite{rmia} on experiments with the CIFAR-10 dataset \cite{cifar10}.  The CIFAR-10 dataset is a widely used benchmark in machine learning, consisting of 60,000 32x32 color images in 10 distinct classes designed for evaluating image classification algorithms.  $50$k of the datapoints are part of the trainig set, with the remaining $10$K as part of the test set. 

\subsubsection{Likelihood Ratio Attack.}
The likelihood Ratio Attack (LiRA) \cite{lira} performs a statistical hypothesis test to determine the membership of a sample in the training set. It operates by training shadow models to estimate the likelihood of a model’s output under both the member and non-member hypotheses. Given $N$ shadow models and a set of canaries, the training datasets are chosen so that each canary is a member for $N/2$ of the models and a non-members for the other $N/2$. We train $8$ models, with one of them being the target model and the other $7$ being the shadow models. Each models is trained with $\approx 25$K samples, and the other $25$K datapoints excluded from training to be used as non-members. 

\begin{table}
\centering
\caption{Accuracy of target and shadow models trained on subsets of CIFAR-10 for LiRA attack}
\label{tab:lira_accuracy}
\begin{tabular}{lcccccccc}
\toprule
 & Target & Shadow 1 & Shadow 2 & Shadow 3 & Shadow 4 & Shadow 5 & Shadow 6 & Shadow 7 \\
\midrule
Train & 96.87 & 96.85 & 96.64 & 96.65 & 96.90 & 96.75 & 96.79 & 96.73 \\
Test  & 84.77 & 84.81 & 84.59 & 84.75 & 85.34 & 84.59 & 84.40 & 85.18 \\
\bottomrule
\end{tabular}
\end{table}

 For each model, we train a ResNet architecture \cite{he2016deep} for $50$ epochs. Final accuracy is shown in Table~\ref{tab:lira_accuracy}. From each model, we extract logits, which follow a Gaussian distribution, for all $50$k datapoints. If $M(x)$ is the vector of predictions of model $M$ on sample $x$ with label $y$ the logits are computed as 
 \begin{align*}
     \phi_M(x, y) = \log\Big(\frac{M(x)[y]}{1-M(x)[y]}\Big).
 \end{align*}
 The adversary uses the logits to compute per-sample scores and decide on the membership status of each sample. Let $M^*$ be the target model. For a given sample $(x, y)$, let $\mathcal{M}_{\mathrm{in}}$ be the set of models trained with $(x, y)$, excluding $M^*$, and let $\mathcal{M}_{\mathrm{out}}$ be the set of models trained without $(x, y)$, excluding $M^*$. The adversary fits Guassian distributions to the logits obtained from $\mathcal{M}_{\mathrm{in}}$ and $\mathcal{M}_{\mathrm{out}}$, and computes its confidence that $(x,y)$ was used in training of $M^*$ based on the logit $\phi_{M^*}(x, y)$.

 There are two variants of LiRA. We have just described the ``online'' version where the adversary trains shadow models both including and excluding the canaries from the training set. In the offline version, which is less resource intensive, the adversary only trains ``out'' models that exclude canaries. It obtains its score with a one-sided hypothesis test, comparing  $\phi_{M^*}(x, y)$ to the Guassian distribution fitted on logits from $\mathcal{M}_{\mathrm{out}}$ models.

\begin{figure}
    \centering
    \begin{subfigure}[b]{0.45\textwidth}
        \centering
        \includegraphics[width=\linewidth]{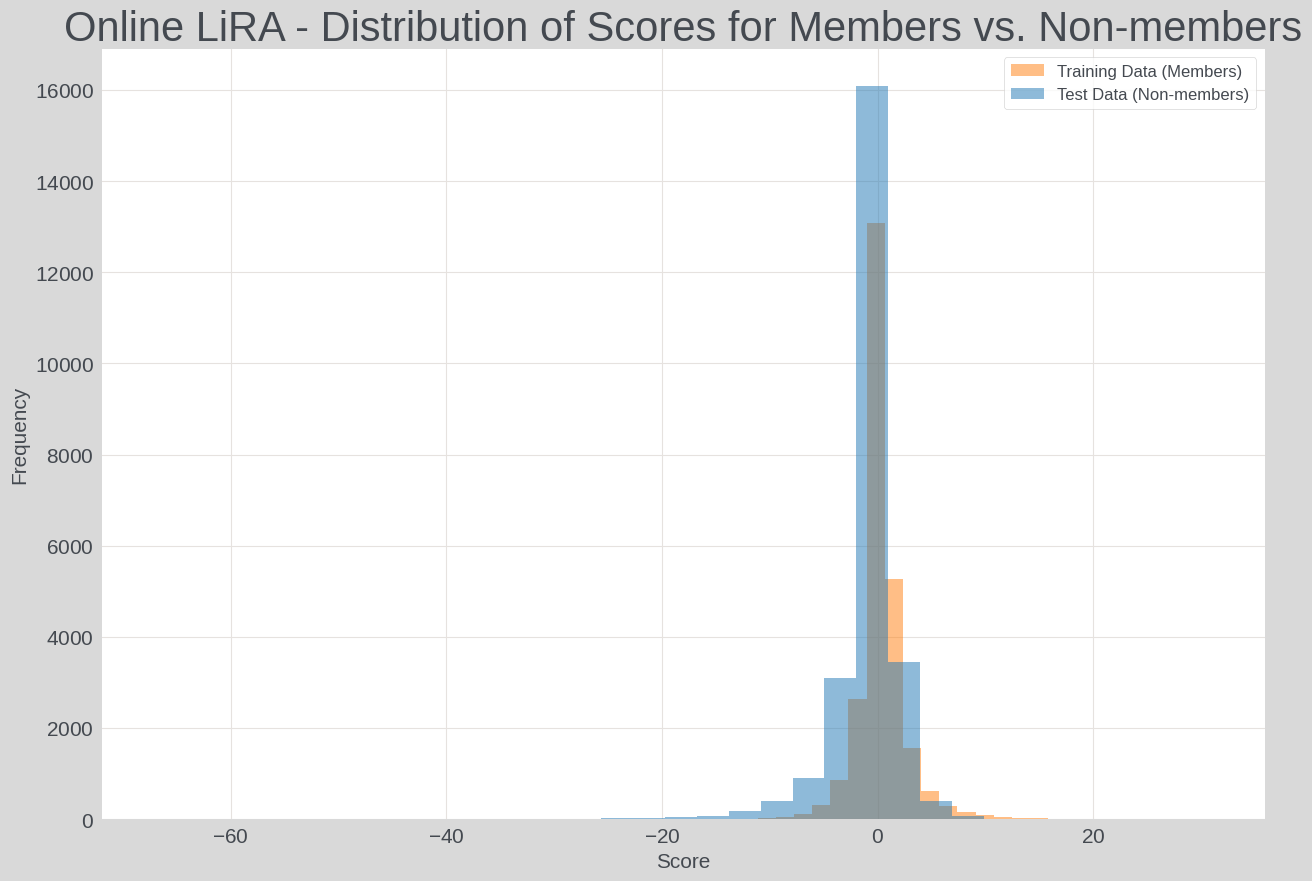}
        \caption{Online LiRA: train shadow models including and excluding canaries.}
        \label{fig:online_lira_scores}
    \end{subfigure}
    \hfill
    \begin{subfigure}[b]{0.45\textwidth}
        \centering
        \includegraphics[width=\linewidth]{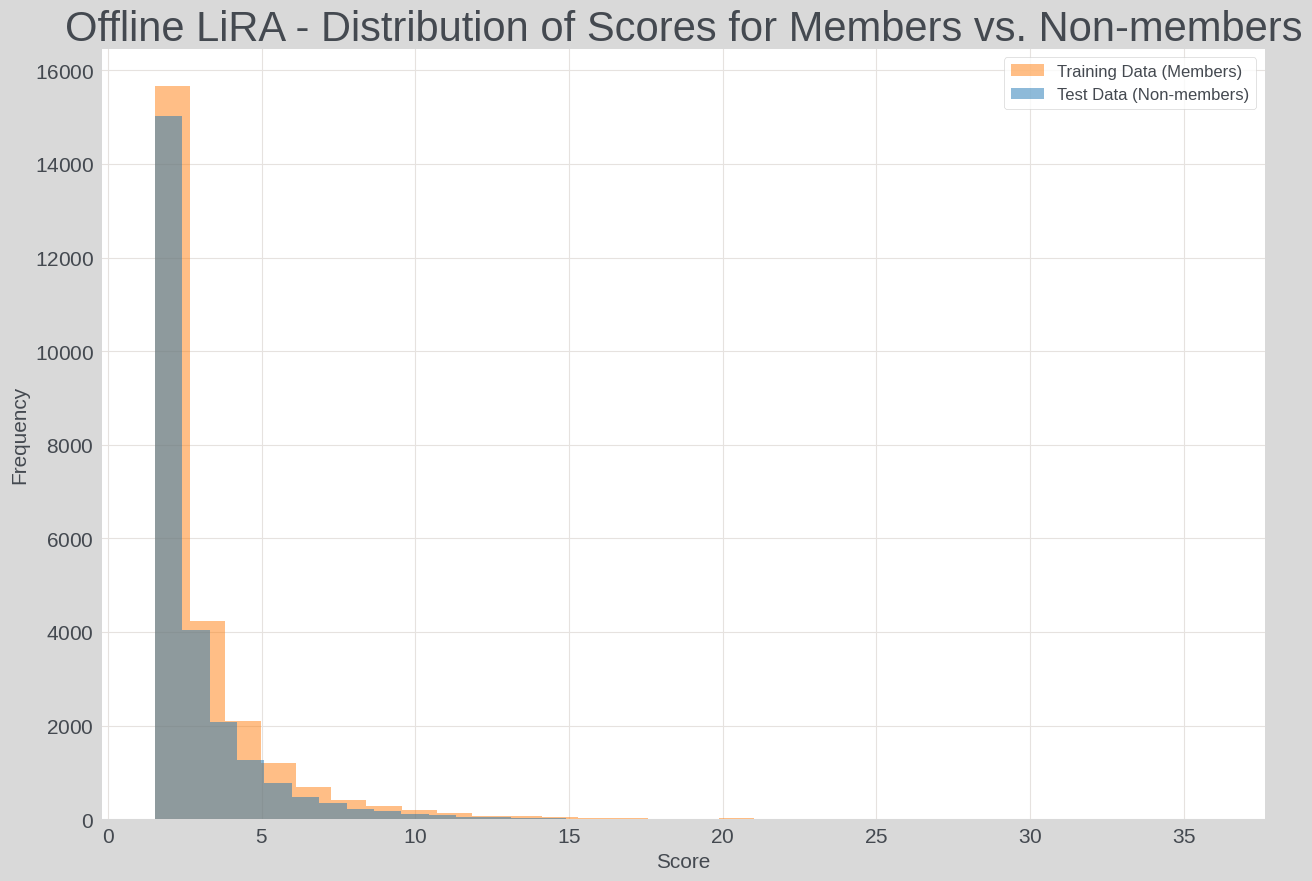}
        \caption{Offline LiRA: only train shadow models excluding canaries}
        \label{fig:offline_lira_scores}
    \end{subfigure}
    \caption{Histogram of per-example scores computed by adversary for likelihood ratio attack (LiRA).}
    \label{fig:lira_scores}
\end{figure}

\begin{figure}
    \centering
    \begin{subfigure}[b]{0.45\textwidth}
        \centering
        \includegraphics[width=\linewidth]{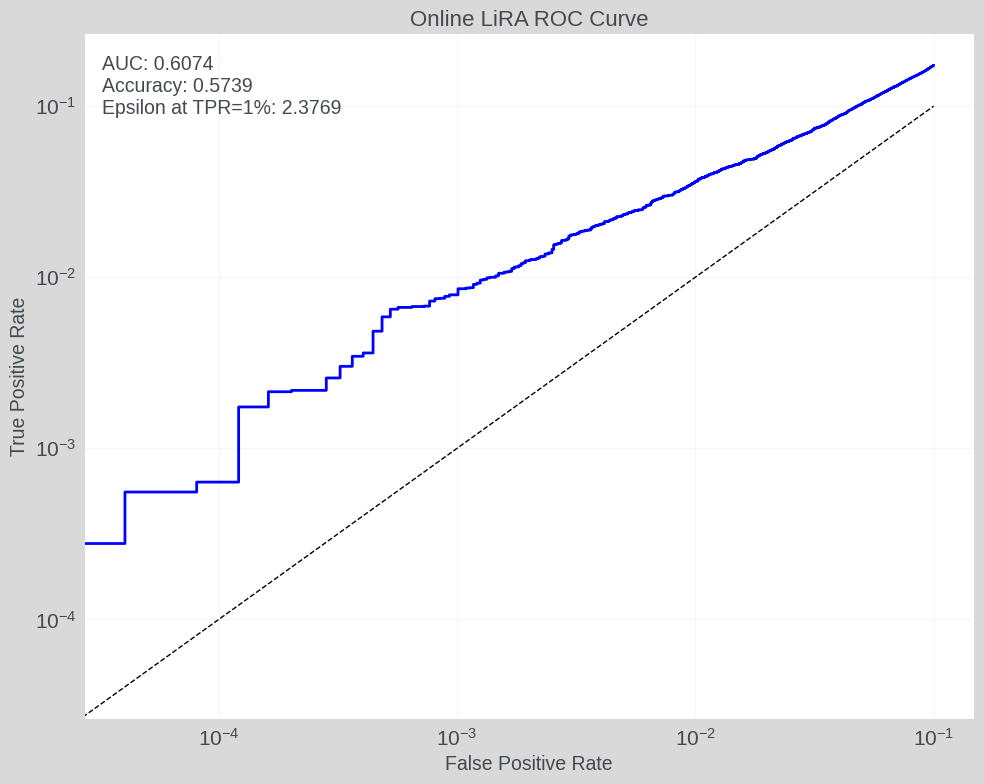}
        \caption{Online LiRA}
        \label{fig:online_lira_roc}
    \end{subfigure}
    \hfill
    \begin{subfigure}[b]{0.45\textwidth}
        \centering
        \includegraphics[width=\linewidth]{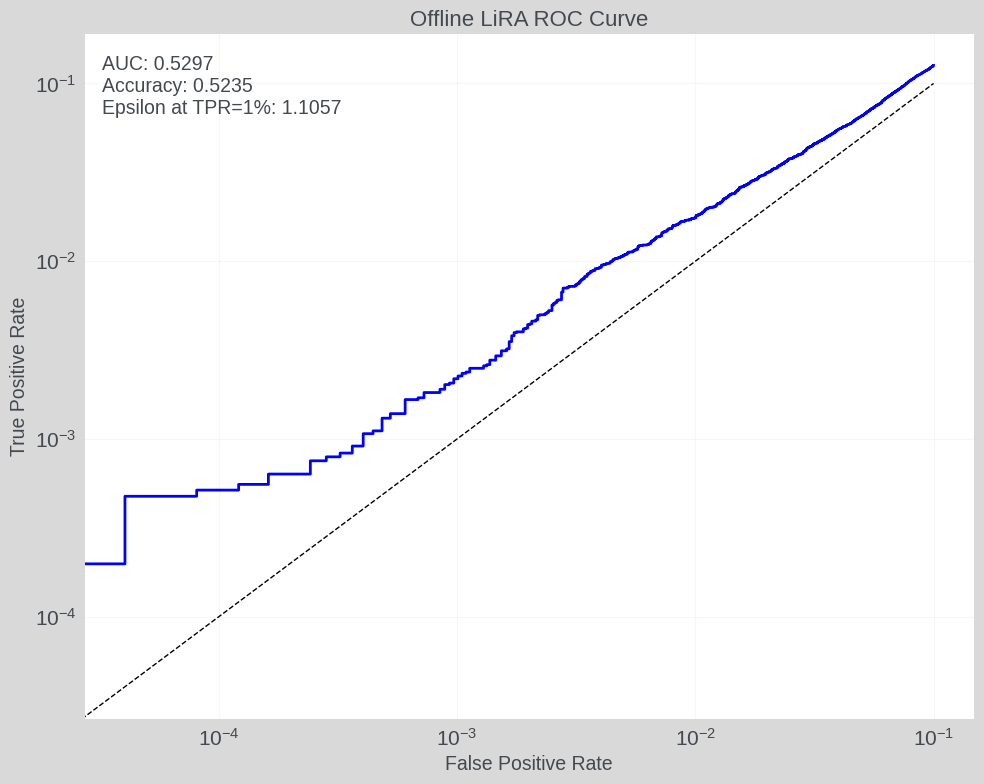}
        \caption{Offline LiRA}
        \label{fig:offline_lira_roc}
    \end{subfigure}
    \caption{ROC curve of adversary, zooming in at low false positive rates. Training members are the ``positive class''}. 
    \label{fig:lira_roc}
\end{figure}

 In Fig.~\ref{fig:lira_scores}, we show the distribution of scores computed by the adversary for the two versions of the attack. We see a shift between the distribution of training members and non-members. In the online LiRA, the adversary can leverage both tails of the score distribution to decide on membership, making it more powerful. Specifically, the adversary uses a score threshold $\tau$ and guesses ``member'' if the per-sample score is greater than $\tau$ and ``non-member'' otherwise. In Fig.~\ref{fig:lira_roc} we show the ROC of the adversary (sweeping over all score thresholds). In the online LiRA the adversary achieves significantly higher AUC (0.6074 for ``online'' versus 0.5297 for ``offline''). 
 
Finally, we compute the empirical epsilon lower bound achieved by each version of LiRA by choosing a score threshold $\tau$ such that the adversary has true positive rate $1\%$, i.e. the adversary can extract $1\%$ of training-set members from the attack. We run $10^3$ rounds of bootstrap to compute confidence intervals on AUC and empirical epsilon (Table~\ref{tab:lira_results}). The online LiRA achieves an empirical epsilon of $2.37$ at $95\%$ confidence compared to $1.11$ for offline LiRA. Ovearll, both models show high levels of memorization, with online LiRA having superior attack performance. 

\begin{table}
\centering
\caption{Results of likelihood ratio attack (LiRA) on CIFAR-10 with $95\%$ CI}
\label{tab:lira_results}
\begin{tabular}{lccc}
\toprule
 Attack &   Epsilon at $1\%$ TPR & AUC & Accuracy \\
\midrule
Online & (1.6004, 2.3769) & 0.6074 (0.6026, 0.6124) & 0.5739 (0.5701, 0.5776)
  \\
Offline  & (0.5637, 1.1057) & 0.5297 (0.5246, 0.5345) &0.5235 (0.5192, 0.5276) \\
\bottomrule
\end{tabular}
\end{table}

\subsubsection{Robust Membership Inference Attack.} The robust membership inference attack (RMIA) is designed to achieve higher power at lower computational resources. It outperforms SOTA attacks such as LiRA \cite{lira}, when limiting these attacks to only a few shadow models (e.g, 1 or 2). In LiRA, we estimate $\frac{\Pr[M \mid x \in D]}{\Pr[M \mid x \notin D}$, where $D$ is the training dataset for $M$, thus comparing two worlds where $x$ is included and excluded from the training set. This test can be seen as averaging over cases where $x$ is removed from the training set and replaced by a population sample $z$:
\begin{align*}
\frac{\Pr[M \mid x \in D]}{\Pr[M \mid x \notin D]} = \mathbb{E}_{z}\Big[ \frac{\Pr[M \mid x \in D]}{\Pr[M \mid x \notin D, z \in D} \Big]     
\end{align*}
Instead of averaging over all population samples $z$, the RMIA makes pairwise comparisons between $x$ and $z$ for many samples $z$ by computing $L(x, z) = \frac{\Pr[M \mid x \in D]}{\Pr[M \mid x \notin D, z \in D]}$.  The value $L(x, z)$ is estimated using a few shadow models, many samples $z$ from the population, and the predictions of these models. We refer to \cite{rmia} for more details. Finally, the score $s(x)$ is computed as the fraction of samples $z$ such that $L(x, z) \geq \gamma$ for some hyper-parameter $\gamma$. 

We run the RMIA on the CIFAR-10 dataset by training one target model and $3$ shadow models. Similarly to LiRA, each model is trained with $25$K datapoints selected at random from the training set, and excluding the remaining $25$K. The ``population'' samples $z$ are sampled without replacement from the training set consisting of $10$K samples. 

\begin{figure}
    \centering
    \begin{subfigure}[b]{0.45\textwidth}
        \centering
        \includegraphics[width=\linewidth]{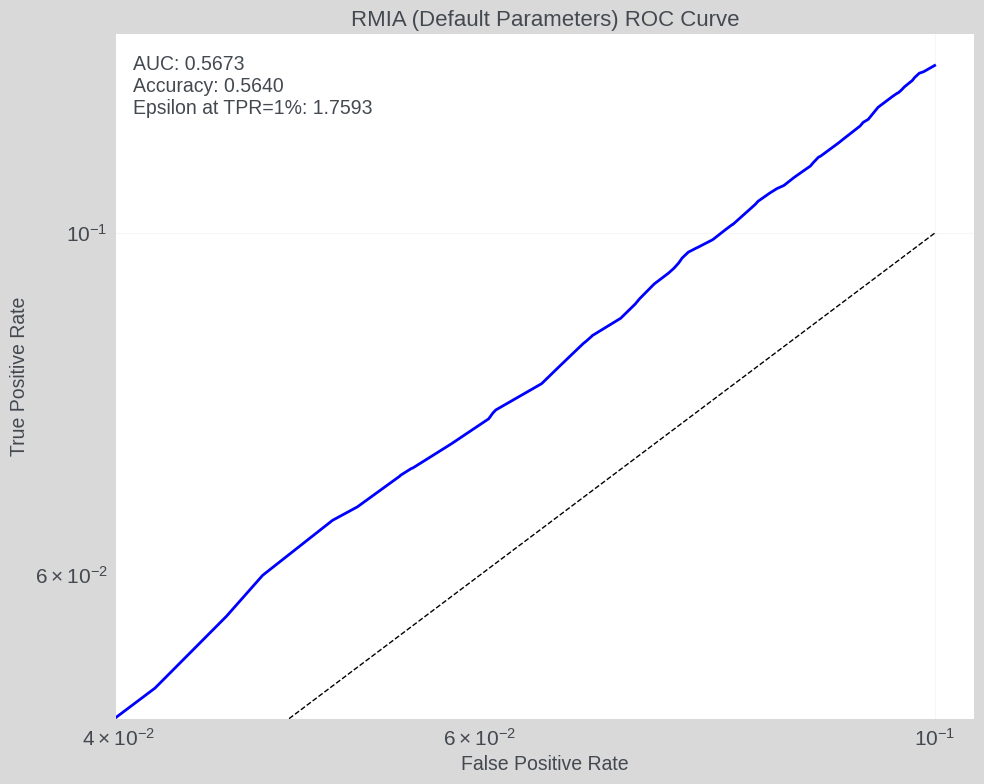}
        \caption{Default $\alpha=0.3$}
        \label{fig:rmia_roc_fixed}
    \end{subfigure}
    \hfill
    \begin{subfigure}[b]{0.45\textwidth}
        \centering
        \includegraphics[width=\linewidth]{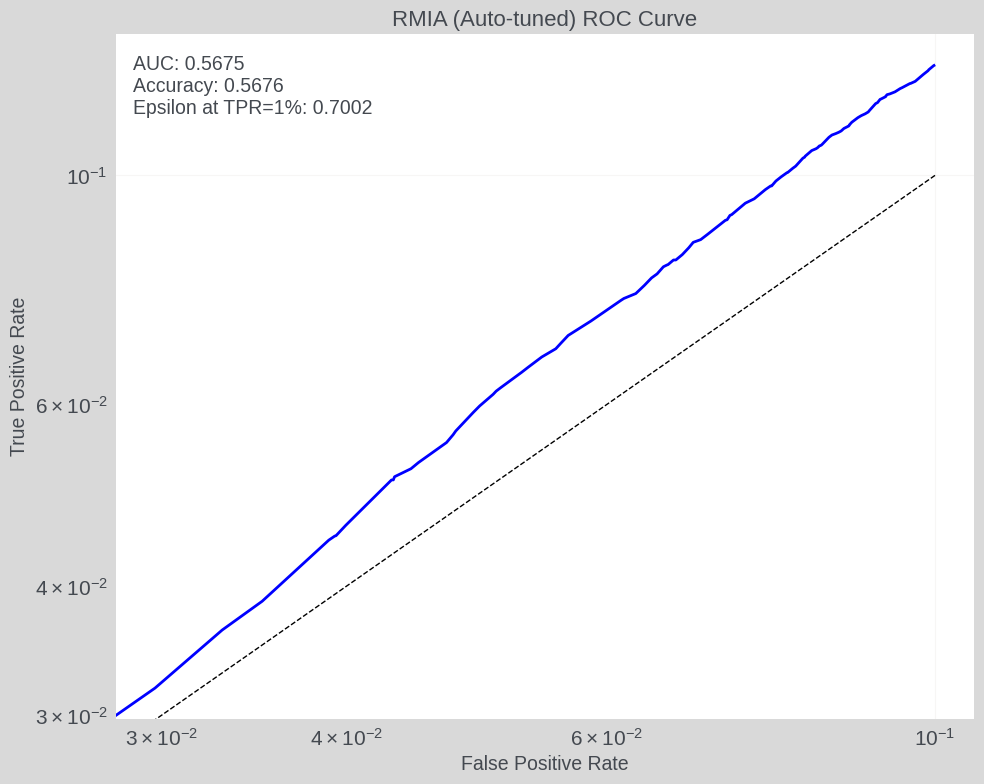}
        \caption{Auto-tuned $\alpha$}
        \label{fig:rmia_roc_tuned}
    \end{subfigure}
    \caption{ROC curve of adversary for RMIA, zooming in at low false positive rates. Training members are the ``positive class''. The parameter $\alpha$ is the gap in predictions for a datapoint between a model including and excluding that datapoint.}
    \label{fig:rmia_roc}
\end{figure}

\begin{figure}
    \centering
    \begin{subfigure}[b]{0.49\textwidth}
        \centering
        \includegraphics[width=\linewidth]{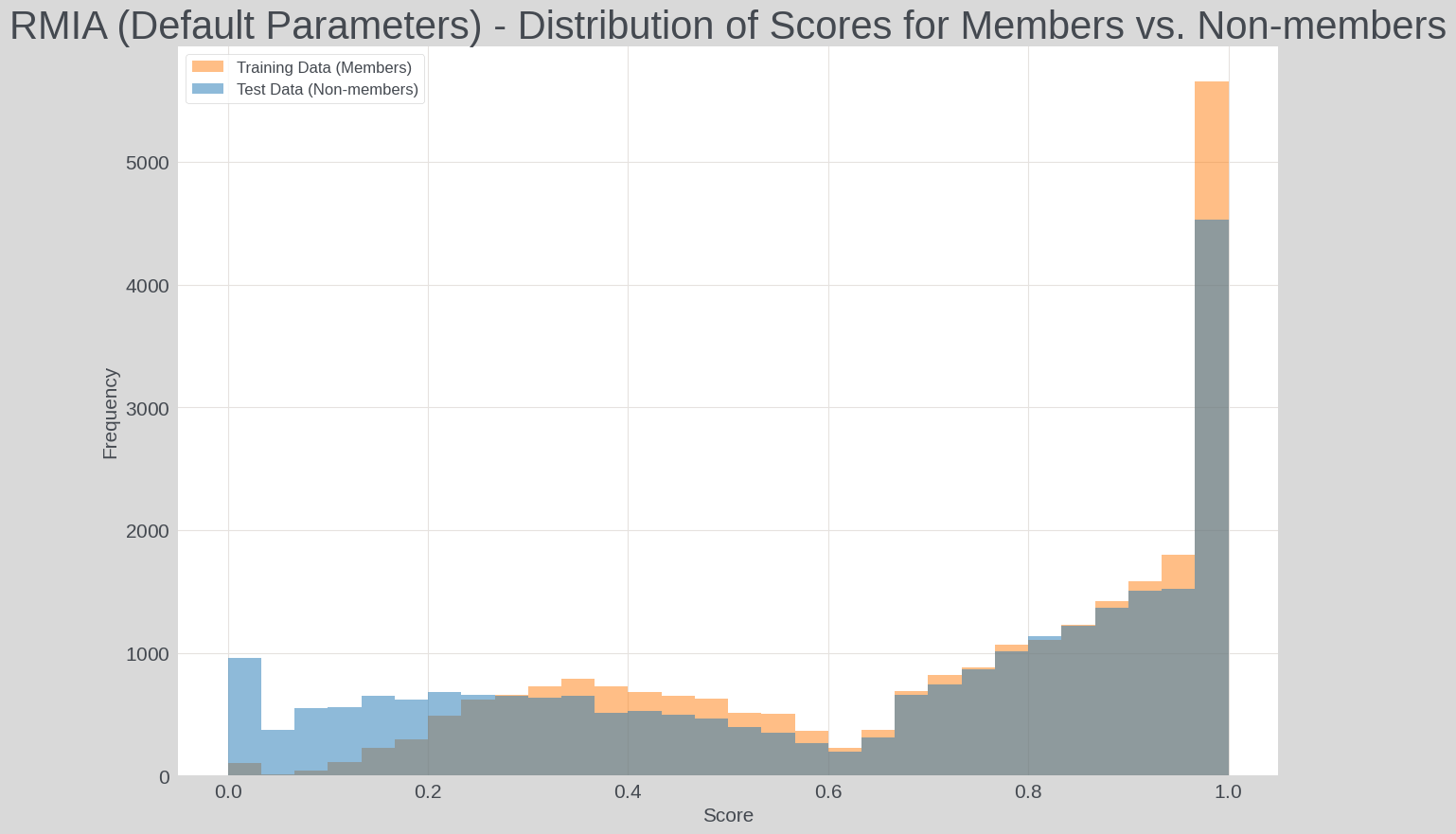}
        \caption{Default $\alpha=0.3$.}
        \label{fig:rmia_scores_fixed}
    \end{subfigure}
    \hfill
    \begin{subfigure}[b]{0.44\textwidth}
        \centering
        \includegraphics[width=\linewidth]{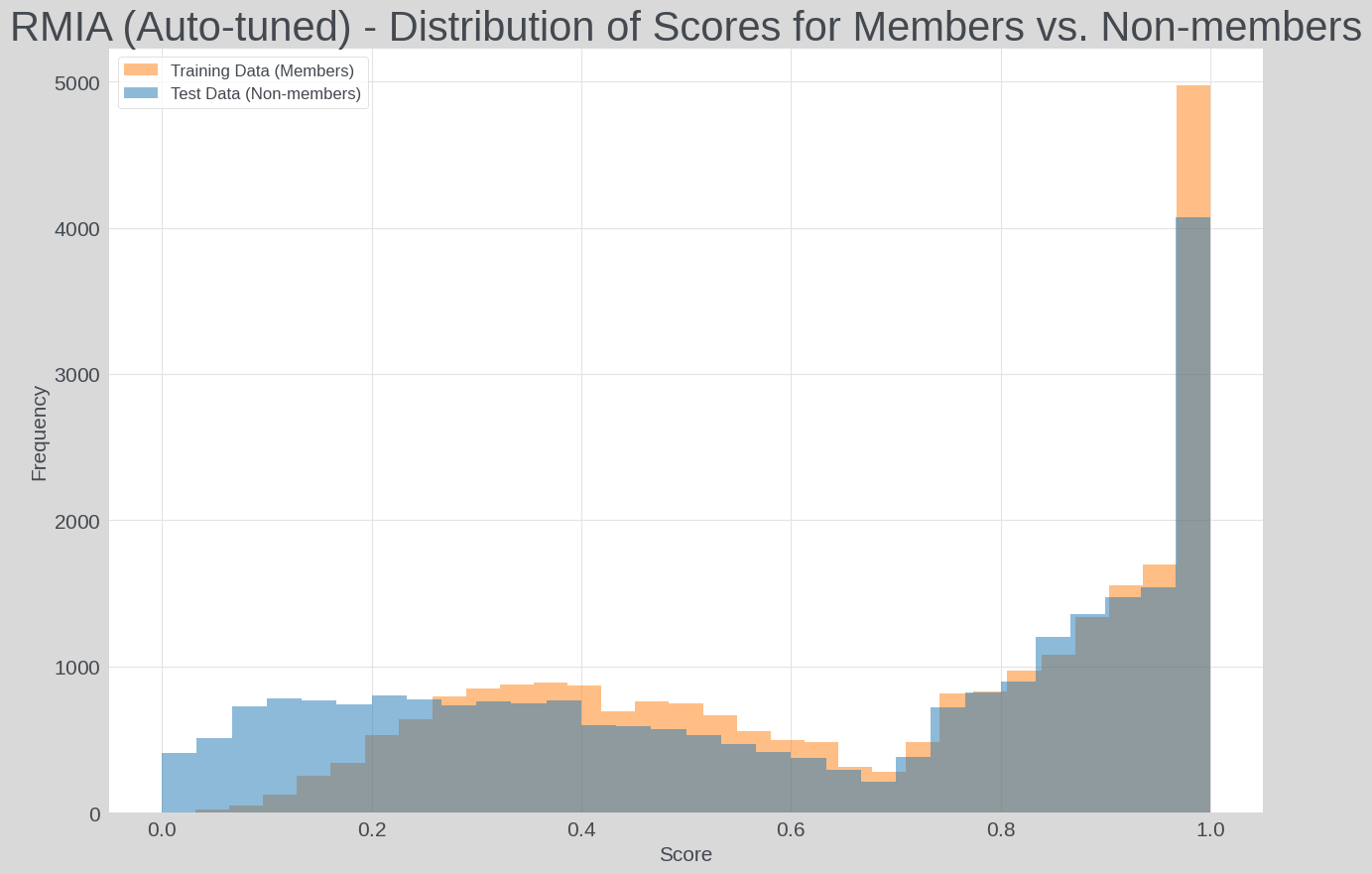}
        \caption{Auto-tuned $\alpha$}
        \label{fig:rmia_scores_tuned}
    \end{subfigure}
    \caption{Histogram of per-example scores computed by adversary for Robust MIA. The parameter $\alpha$ is the gap in predictions for a datapoint between a model including and excluding that datapoint.}
    \label{fig:rmia_scores}
\end{figure}

A key hyper-parameter of the attack is $\alpha$, an estimate of the gap in average prediction between models trained with and without a given sample. Such an estimate is necessary because the attack only trains shadow models that exclude population samples $z$ and does not train models including them. We compare an attack with a default value of $\alpha=0.3$ and another attack which tunes $\alpha$ automatically. The attack which auto-tunes $\alpha$ shows better separation of non-members and non-members on the left-hand tail of the score distribution (Fig.~\ref{fig:rmia_scores}), but the two attacks achieve similar AUC (Fig.~\ref{fig:rmia_roc}).

Note that RMIA achieves a higher AUC of $0.5673$ (Table~\ref{tab:rmia_results}) compared to the AUC of $0.5297$ of offline LiRA (Table~\ref{tab:lira_results}) while using the same number of shadow models.

\begin{table}
\centering
\caption{Results of robust membership inference attack (RMIA) on CIFAR-10 with $95\%$ CI}
\label{tab:rmia_results}
\begin{tabular}{lccc}
\toprule
 Attack &   Epsilon at $1\%$ TPR & AUC & Accuracy \\
\midrule
Default $\alpha$ & (-0.7524, 1.7593) & 0.5673 (0.5625, 0.5723)
 & 0.5640 (0.5611, 0.5671) \\
Auto-tuned $\alpha$   & (-0.6961, 0.7002) & 0.5675 (0.5626, 0.5725) & 0.5676 (0.5646, 0.5709) \\
\bottomrule
\end{tabular}
\end{table}

\subsection{Auditing Analyses}

PrivacyGuard supports two methods for differential privacy auditing that translate the output of an attack into a lower bound on $\epsilon$ in the $(\epsilon, \delta)$-DP definition (Definition~\ref{def:dp}). 

 \begin{definition}[\cite{DMNS06}]
 \label{def:dp}
     A mechanism $M$ is $(\epsilon, \delta)$-differentially private if for all datasets $D, D'$ that differ only at one sample, and all outputs $S$ of the mechanism
     \begin{align}
         \Pr[M(D) \in S] \leq e^\epsilon \Pr[M(D') \in S] + \delta.
      \end{align}
 \end{definition}
The first method obtains confidence intervals on the empirical epsilon lower bound using bootstrapping on the canary set. This method assumes that the canaries used for the attack produce independent results. However, in many attacks, the model is trained on all canaries simultaneously (these are known as one-run attacks) which breaks the independence assumption. To properly account for this dependence, we also implement the auditing method of \cite{mahloujifar2025auditing}. It provides the tightest auditing result in the literature, by targeting a more expressive notion of DP called $f$-DP \cite{dong2019gaussian}. 

In this section, we describe the two analysis methods. We have demonstrated the bootstrap-based analysis with our attacks on CIFAR-10. Here we demonstrate the $f$-DP auditing analysis with synthetic data.  

\subsubsection{Bootstrap-based analysis. } Let $k$ be the number of boostrap runs, which is a user-provided argument. Let $m$ be the number of samples/canaries used in the attack. We repeat the following for all $k$ rounds. We subsample \textit{without replacement} $m$ samples with their associated attack scores. For each round, we obtain adversarial success metrics: empirical epsilon values at different score thresholds, AUC, and accuracy. Finally, we compute $95\%$ confidence intervals (across the $k$ rounds) for each success metric. 

We compute AUC by considering all scores and membership values. We compute empirical epsilons by computing the TPR/FPR and TNR/FNR ratio at different score thresholds. More formally, given a score threshold $\tau$ and score $s(x)$ for sample $x$, the adversary guesses ``member'' if $s(x) \geq \tau$ and non-member otherwise. Let the set of canaries be $\{x_i \mid i\in [m]\}$, $b(x_i) \in \{0,1\}$ is the membership status of $x_i$ ad $s(x_i)$ is the score computed by the attacker. For the threshold $\tau$ the true positive rate $\mathrm{TPR}_{\tau}$ is 
\begin{align*}
    \mathrm{TPR}_{\tau} = \frac{\sum_{i \in [m]} \mathbf{1}[s(x_i) \geq \tau]\cdot \mathbf{1}[b(x_i) = 1]}{\sum_{i \in [m]} \mathbf{1}[b(x_i) = 1]} 
\end{align*}
We can compute $\mathrm{TNR}_{\tau}$, $\mathrm{FPR}_{\tau}$, and $\mathrm{FNR}_{\tau}$ similarly. By $(\epsilon, \delta)$-differential privacy \cite{kairouz2015composition}:
\begin{align}
\label{eq:epsilon}
    \epsilon \geq \max\Big\{  \ln\Big(
    \frac{\mathrm{TPR}_{\tau} - \delta}{\mathrm{FPR}_{\tau}} \Big), 
    \ln\Big(\frac{\mathrm{TNR}_{\tau} - \delta}{\mathrm{FNR}_{\tau}}\Big)
    \Big\}
\end{align}
We iterate over score thresholds $\tau$ from the set:
\begin{align*}
    T = \{ \tau \text{ s.t. } \mathrm{Error}_{\tau}  = p \mid p \in (0.01, 0.02, \dots, 0.99), \mathrm{Error} \in \{\mathrm{TPR, FPR, TNR, FNR}\} \} 
\end{align*}
Then for each $\tau$ in $T$ we compute $\epsilon_\tau$ according to the left hand side of (\ref{eq:epsilon}). This is repeated for $k$ rounds, so that we can compute confidence intervals on the empirical epsilon obtained for each $\tau \in T$. The final ``empirical epsilon`` is the maximum $95\%$ CI upper bound over all $\tau \in T$. 

\subsubsection{$f$-DP auditing \cite{mahloujifar2025auditing}.}
The work of \cite{mahloujifar2025auditing} builds off of \cite{steinke2024privacy}, which consider the accuracy of the adversary that guesses membership status. These works provide an upper bound on the tail of this random variable. \cite{steinke2024privacy} show that when $\delta=0$, the tail of the random variable is bounded by a binomial distribution $\mathsf{Binomial}(m, p)$, where $p = \frac{e^\epsilon}{e^\epsilon+1}$ and $m$ is the number of canaries. When $\delta > 0$, the bound has an additive error of $O(m \cdot \delta)$. When $\delta > 0$, their bound is not tight for simple mechanisms which add Gaussian noise. \cite{mahloujifar2025auditing} provide tighter bounds by auditing the $f$-DP curve of the algorithm (\cite{dong2019gaussian}), which allows for a  more fine-grained privacy accounting then Definition~\ref{def:dp}. The empirical lower bounds on $f$-DP are then translated into lower bounds on the traditional $(\epsilon, \delta)$-DP definition.

\begin{figure}
    \centering
    \begin{subfigure}[b]{0.45\textwidth}
        \centering
        \includegraphics[width=\linewidth]{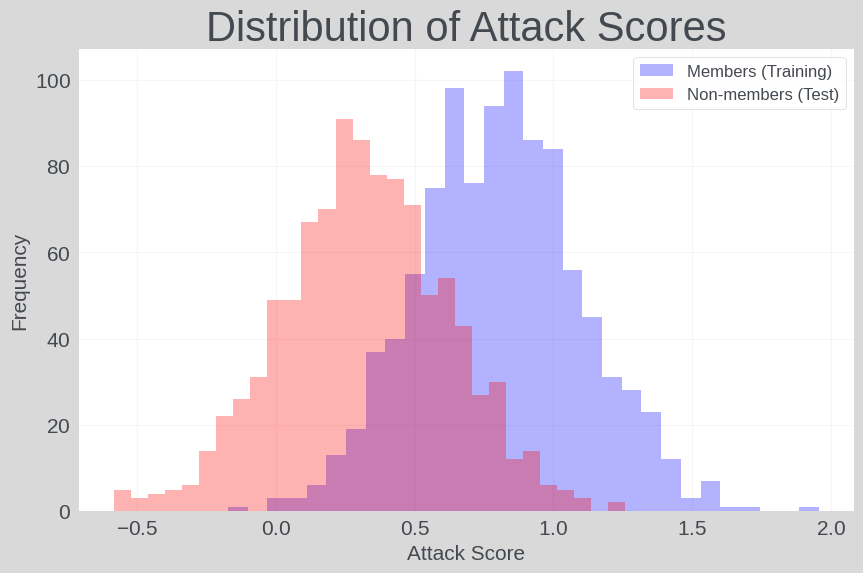}
        \caption{Distribution of scores}
        \label{fig:fpd_scores}
    \end{subfigure}
    \hfill
    \begin{subfigure}[b]{0.4\textwidth}
        \centering
        \includegraphics[width=\linewidth]{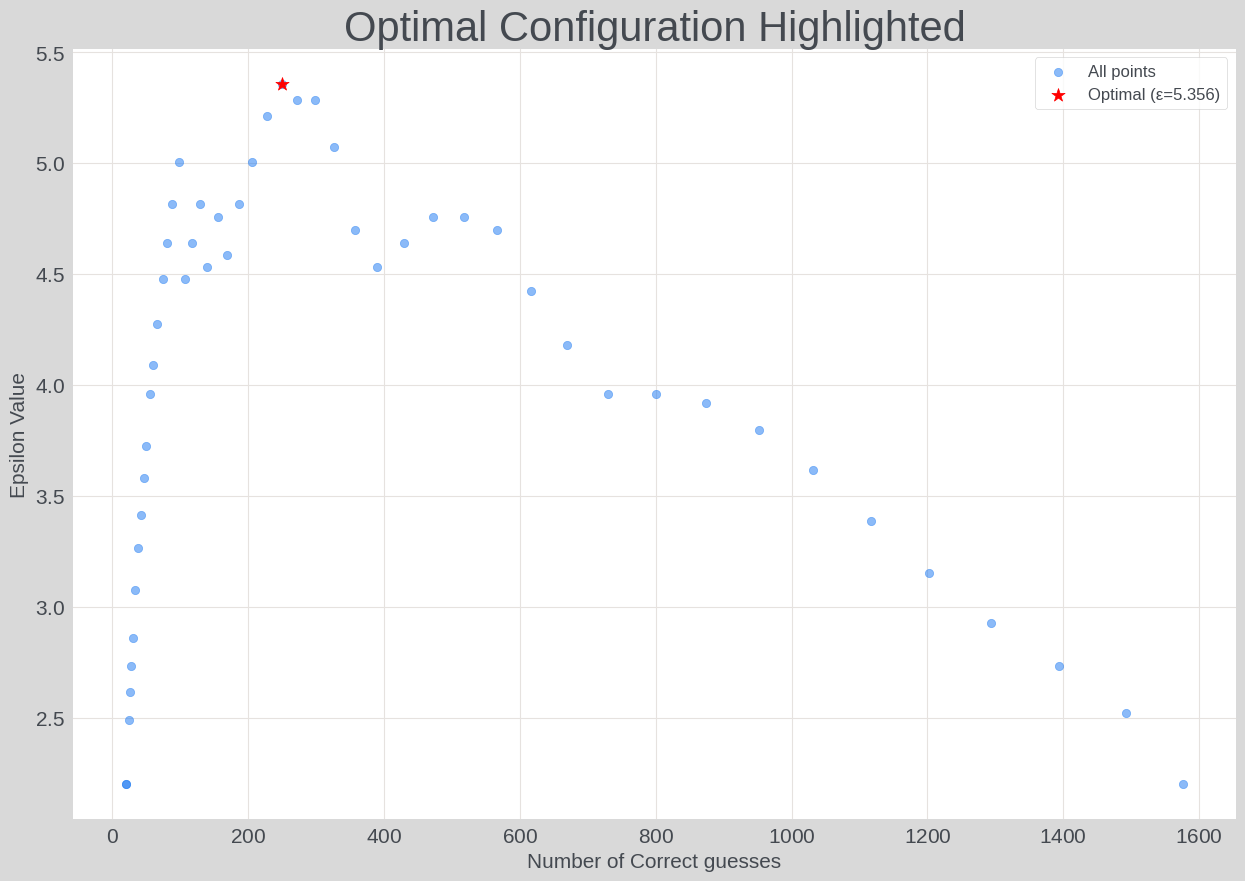}
        \caption{Empirical epsilon for different number of correct guesses}
        \label{fig:fpd_epsilon}
    \end{subfigure}
    \caption{Using $f$-DP auditing \cite{mahloujifar2025auditing} with synthetic scores to obtain empirical epsilon at various number of correct guesses}
    \label{fig:fdp}
\end{figure}

We provide a tutorial that uses synthetic data to demonstrate the use of $f$-DP analysis to obtain an empirical epsilon value. We obtain synthetic scores for members and non-members distributed as two shifted Gaussian distributions (Fig.~\ref{fig:fpd_scores}). There are $m = 2$K canaries in total, with $1$K for each class. The $\mathsf{FDPAnalysisNode}$ class takes as inputs: the number of canaries $m$ , the number of correct guesses by the adversary $c$, and the total number of guesses $\hat{c}$. The adversary can choose to abstain on samples on which it is least confidence, thus $\hat{c} \leq m$. By allowing the adversary to abstain, we can obtain stronger bounds on $\epsilon$, however the number of abstentions cannot be so large that the confidence interval on the empirical epsilon is very large. We iterate over $\hat{c} \in \{10, \dots, \log_{10}(m)\}$. The adversary can guess on either $\hat{c}$ samples with highest scores (by guessing member for all of them), or guess member for $\hat{c}/2$ highest scores and non-members for $\hat{c}/2$ lowest scores. In Fig.~\ref{fig:fpd_epsilon}, we plot the empirical epsilon achieved for each $\hat{c}$ and guessing strategy (one-sided versus two-sided). The optimal configuration is the one which gives the highest empirical epsilon ($\epsilon = 5.356)$, achieved with two-sided guessing and $249$ correct guesses. 

\subsection{Probabilistic Extraction on Language Models}

This section specifies the methodological details of the extraction attacks supported by PrivacyGuard to study memorization in large language models (LLMs). We adopt the notation and framing introduced in \cite{hayes-etal-2025-measuring}, and we connect it to prior work on extraction attacks \cite{carlini2022,nasr2023}. We consider two complementary settings:
\begin{itemize}
  \item \textbf{Discoverable Extraction (completion-based):} Given a training example $(x|z)$, it attempts to elicit the target string $z$ via completions under a specified prefix prompt ($x$) and decoding policy.
  \item \textbf{Probabilistic Discoverable Extraction:} Given a training example $(x|z)$, it quantifies the probability of extracting a fixed target string $z$ under a specified prompt ($x$) for (potentially) multiple decoding policies. This yields an empirical extraction probability $p_z$ and an interpretable two-parameter abstraction $(n,p)$, which captures the probability of outputting $z$ in at least once in $n$ attempts.
\end{itemize}

\paragraph{Notation.}
Let $f_{\theta}$ denote an autoregressive language model with parameters $\theta$, vocabulary $\Sigma$, trained on dataset $D$. We let $(x|z) \in D$ denote a canonical training example with $z\in\Sigma$ being the canonical \emph{target} and $x\in\Sigma$ being the canonical \emph{attack prompt} (prefix). A \emph{sampling scheme}, parameterized by $\phi$ (e.g., greedy, temperature $T$, top-$p$, top-$k$), induces a sampler $g_{\phi}$ that transforms next-token probabilities into an actual draw. A single attempt with prompt $x$ produces a random completion $Y$ via the composition $Y \sim (g_{\phi}\circ f_{\theta})(\cdot \mid x)$. We write $\texttt{match}(Y,z)\in\{0,1\}$ for a fixed normalization/matching predicate that declares success. Randomness arises solely from $g_{\phi}$ (e.g., its seed). A query budget $N$ specifies the number of independent attempts with the same $(x,\phi)$.

\subsection*{Discoverable Extraction (completion-based)}
Operationally, we fix $x$ and $\phi$ and extract (draw) $Y\sim(g_{\phi}\circ f_{\theta})(\cdot\mid x)$ using a single attempt. We test the similarity of the extraction $Y$ with the target $z$ using a matching function $\texttt{match}(Y,z)=1$. Under greedy decoding, the extraction $Y$ is deterministic and under a stochastic $\phi$, the extraction $Y$ is a random draw from the tilted token probability distribution defined by the sampling scheme, using a given seed.

\begin{definition}[Discoverable Extraction]
Given a training example $(x|z) \in D$ and a fixed decoding scheme $\phi$, we say that $z$ is \textit{discoverably extractable} using $x$ iff
$\texttt{match}\big((g_{\phi}\circ f_{\theta})(x),\,z\big)=1$, for a suitably defined matching function \texttt{match}.
\end{definition}

\paragraph{The matching function $\texttt{match}(\cdot,\cdot)$ can be set as:}
\begin{itemize}
  \item \emph{Exact equality:} $\texttt{match}(Y,z)=\mathbf{1}\{\,\texttt{norm}(Y)=\texttt{norm}(z)\,\}$, where $\texttt{norm}$ applies Unicode/whitespace normalization.
  \item \emph{Substring inclusion:} $\texttt{match}(Y,z)=\mathbf{1}\{\,\texttt{norm}(z)\ \text{is a contiguous substring of}\ \texttt{norm}(Y)\,\}$.
  \item \emph{LCS threshold:} letting $\texttt{LCS}(\cdot,\cdot)$ be the longest common subsequence on tokens, $\texttt{match}(Y,z)=\mathbf{1}\{\,\texttt{LCS}(\texttt{norm}(Y),\texttt{norm}(z)) \ge \tau\,\}$ for a user-chosen threshold $\tau$.
\end{itemize}

Discoverable extraction is the most widely used operationalization of memorization in both industry and academia (e.g., \cite{carlini2022,nasr2023,gemma25,llama24,biderman23}). Numerous variants instantiate different token lengths for the cue $x$ and the target $z$—for example, \cite{biderman23} uses 32 tokens for each, while \cite{carlini2022} explores lengths ranging from 50 to 500 tokens. In practice, discoverable extraction is especially well-suited to \emph{greedy} decoding, where the decoding rule deterministically selects the argmax token at each step, making the metric itself deterministic and reproducible for a fixed prompt. By contrast, modern LLM deployments frequently rely on stochastic sampling schemes to promote diversity and creativity: \emph{temperature} sampling rescales logits by $1/T$ before softmax; \emph{top-$k$} sampling restricts draws to the $k$ highest-probability tokens; and \emph{top-$p$} (nucleus) sampling restricts to the smallest set of tokens whose cumulative probability exceeds $p$, renormalizing within that set. Because these schemes introduce randomness, real users may sample multiple times for the same $x$, implicitly performing repeated trials. This motivates the \emph{probabilistic discoverable extraction} view, which treats the single-attempt exact-match probability for $z$ under a fixed sampling scheme as fundamental, and then reasons about the probability of at-least-one success across multiple (independent) attempts.

\subsection*{Probabilistic Discoverable Extraction (exact-match only)}
Rather than a one-shot, yes/no test, as we do in discoverable extraction, it is often more informative to ask \emph{how many} sequences a user must generate before a target example is likely to be extracted under a chosen sampling scheme. The $(n,p)$ notion of discoverable extraction formalizes this idea: it characterizes the capability of a regular user who can query the model $n$ times and extract the target \emph{verbatim} with probability at least $p$. Under non-deterministic decoding (e.g., temperature, top-$k$, top-$p$), many targets will not be guaranteed ($p=1$) with a single query, yet may be produced at least once when the user is permitted multiple independent attempts ($n>1$). This yields a \emph{continuous} view of extraction risk: instead of a brittle, one-shot outcome, $(n,p)$ quantifies how the likelihood of exact extraction scales with additional opportunities to sample, providing a potentially more meaningful estimate of memorization risk than can be gleaned from a single trial.

\begin{definition}[$(n,p)$-discoverable extraction]
Given a training example $(x|z) \in D$ and a fixed decoding scheme $\phi$, we say that $z$ (with $n_z$ tokens) is $(n,p)$\emph{-discoverably extractable} (with respect to $f_{\theta}$ and $g_{\phi}$  using $x$ iff
\[
\Pr\!\left( \bigcup_{w\in[n]} \Big\{\, (g_{\phi}\circ f_{\theta})^{n_z}_{w}(x) \;=\; z \,\Big\} \right) \;\ge\; p,
\]
where $(g_{\phi}\circ f_{\theta})^{n_z}_{w}(x)$ denotes the $w$-th (of $n$) independent execution of the autoregressive process that, starting from the same initial sequence $x$, applies $g_{\phi}\circ f_{\theta}$ to generate and append $n_z$ tokens in sequence.
\end{definition}

Operationally, we take the training data point $(x|z)$ and split it into a known prefix $x$ and a $n_z$-token suffix $z$. Use the prefix as the prompt, $x$, we fix a sampling scheme $g_{\phi}$, and allow a user $n$ independent opportunities to continue the sequence. Each opportunity runs the autoregressive generator $(g_{\phi}\!\circ\! f_{\theta})$ for exactly $n_z$ steps, yielding a candidate continuation from the same starting prefix. If the probability of generating $z$ exactly at least once is larger than $p$, then we say $z$ is $(n,p)$-discoverably extractable. 

In the ideal setting without considering compute requirements, one can use a Monte Carlo approach to estimate $(n,p)$-discoverably extractability by sampling $N$ independent runs of $n$ independent completions each from $(g_{\phi}\!\circ\! f_{\theta})(\cdot\mid x)$ and estimate
$\hat p = \frac{1}{N}\sum_{j=1}^{N}\mathbf{1}\{\bigcup_{w\in[n]}(g_{\phi}\!\circ\! f_{\theta})^{n_z}_{w}(x) = z\}$, and report the sequence as $(n,\hat p)$-discoverably extractable. However, this method of sampling is extremely expensive since it requires $n \times N$ independent generations of length $n_z$ each. Thus, for an efficient and high fidelity approximation, we compute (n,p) using a one-query trick as described in \cite{hayes-etal-2025-measuring} which we describe below.

For fixed $(x,\phi)$, let
\[
p_z \;=\; \Pr\!\big( (g_{\phi}\circ f_{\theta})(x) = z \big),
\]
i.e., the \emph{single-attempt} probability mass that the sampler generates $z$ exactly. 
Because $(g_{\phi}\circ f_{\theta})$ defines an explicit sequence distribution, $p_z$ is computed \emph{exactly} as the product of next-token probabilities along $z$. Writing $z=y_{1:|z|}$ and $h_i = x \oplus y_{<i}$, we have $p_z=\prod_{i=1}^{|z|} q_{\phi}(y_i \mid h_i)$. Here $q_{\phi}(\cdot\mid h)=(g_{\phi}\circ f_{\theta})(\cdot\mid h)$ is the effective next-token distribution under $\phi$. Under greedy, $q_{\phi}$ is a point mass, so $p_z\in\{0,1\}$.

Now, the probability of not generating $z$ in a single draw from the sampling scheme is $1 - p_z$, and the probability of not generating $z$ in $n$ independent draws is $(1 - p_z)^{n}$. Therefore, the example $z$ is $(n,p)$-discoverably extractable for any $n$ and $p$ that satisfy
\[
1 - (1 - p_z)^{n} \;\ge\; p.
\]
Equivalently,
\[
n \;\ge\; \frac{\log(1 - p)}{\log(1 - p_z)}. \tag{2}
\]
In other words, given a fixed $p$ and the single-attempt exact-match probability $p_z$, one can directly solve for $n$ (and vice versa) using the following equations:
\[
\text{(known $n$)}\quad p \;=\; 1-(1-p_z)^n, \qquad
\text{(known $\alpha$)}\quad \widehat{n} \;=\; \frac{\log(1-p)}{\log(1-p_z)}.
\]

\subsubsection{Experiments} 
\paragraph{Models and datasets}
We evaluate a single autoregressive model ($f_{\theta}$) \textbf{Pythia-12B}~\cite{biderman23}.
We consider the \textbf{Enron Emails} dataset with lightweight, public preprocessing to facilitate replication. Raw messages are parsed with \texttt{mailparser}, and we extract the message body. We retain messages whose body contains at least $100$ tokens under the tokenizer $\tau$, then uniformly sample $10{,}000$ examples.

\paragraph{Extraction attack methodology.}
From each document, we choose training examples as $x|z$ by taking the first $100$ eligible tokens of the body and splitting into a $50$-token prefix ($x$) and a $50$-token suffix ($z$). We evaluate:
\begin{itemize}
  \item \textbf{Discoverable extraction (completion-based).} With $x$ and decoding scheme $\phi$, we generate a single completion $Y\sim (g_{\phi}\!\circ\! f_{\theta})(\cdot\mid x)$ and record success under three predicates: (i) \emph{Exact equality} , (ii) \emph{Substring inclusion}, and (iii) \emph{LCS threshold} $\mathrm{LCS}(Y,z)\ge \tau$.
  \item \textbf{Probabilistic discoverable extraction (exact match only).} For the same $(x,\phi)$, we compute the single-attempt exact-match probability
  $p_z=\Pr\!\big((g_{\phi}\!\circ\! f_{\theta})^{50}(x)=z\big)$
  exactly from stepwise probabilities. For a user with $n$ independent attempts, the success probability is summarized as $\alpha@n = 1-(1-p_z)^n$.
\end{itemize}

\paragraph{Sampling schemes.}
Unless otherwise stated, we consider $\phi\in\{\mathrm{greedy},\,\mathrm{top}\text{-}k,\,\mathrm{temperature}\ T\}$ using standard hyperparameter grids (e.g., $k\in\{40\}$, $T = 1.0$). Greedy yields deterministic discoverable extraction.

\begin{table}[h!]
\centering
\large
\begin{tabular}{l|c|c|c|c|c}
\toprule
Dataset & $\phi$ & Inclusion & LCS $(> 0.8)$ & $p_z(> 0.5)$ & $p_z(> 0.01)$\\
\midrule
Enron & greedy  & 0.049 & 0.0645 & 0.045 & 0.045 \\
Enron & top-$k$ & 0.02 & 0.03 & 0.014 & 0.068 \\
\bottomrule
\end{tabular}
\caption{Extraction rates on Enron dataset}
\label{tab:extraction}
\end{table}

\begin{figure}
    \centering\includegraphics[width=0.7\linewidth]{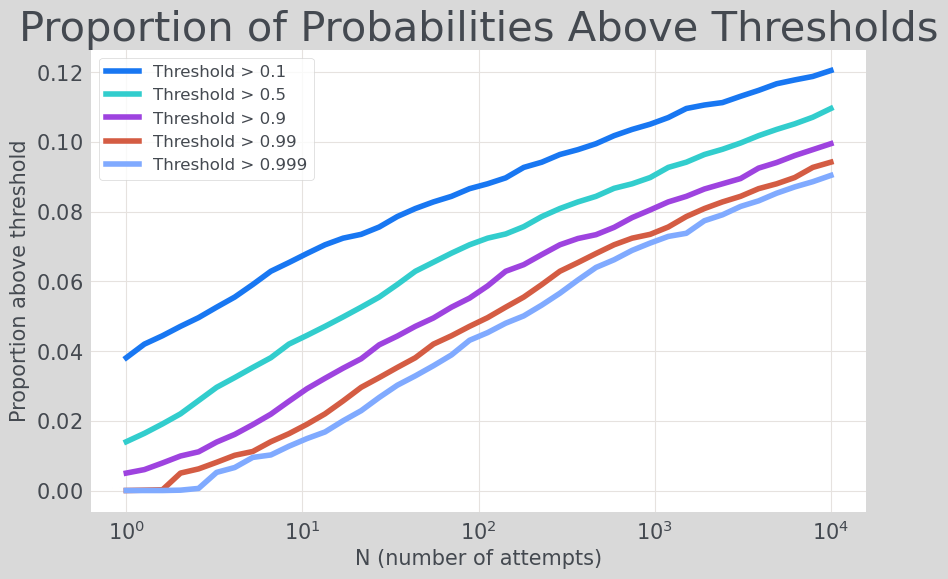}
    \caption{Memorization based on $(n,p)$-discoverable extractability}
    \label{fig:prob_mem}
\end{figure}

\paragraph{Takeaways.} Table~\ref{tab:extraction} reports discoverable and probabilistic discoverable extraction outcomes on the Enron dataset for \textsc{Pythia-12B} under greedy and top-$k{=}40$ decoding. As expected, the approximate token-level LCS predicate yields higher one-shot extraction rates than the substring \emph{Inclusion} predicate, reflecting its greater tolerance to minor edits and boundary effects. For greedy decoding, the computed single-attempt exact-match probabilities are point-mass ($p_z\in\{0,1\}$), so thresholds $p_z>0.5$ and $p_z>0.01$ coincide. In contrast, under top-$k$, thresholding by $p_z$ is informative: the fraction of targets with $p_z>0.01$ exceeds that with $p_z>0.5$, consistent with the stochastic sampler assigning nontrivial but sub-majority mass to some memorized suffixes. Overall, $p_z$-based thresholding provides a more nuanced view of memorization, especially for non-greedy schemes where exact-match success is inherently probabilistic. Figure~\ref{fig:prob_mem} complements this with a practical $(n,p)$ lens, plotting the fraction of targets deemed $(n,p)$-extractable as a function of the number of independent attempts $n$ (x-axis) and several target probabilities $p$ (legend). Curves rise monotonically with $n$ and shift downward for stricter $p$, as anticipated from $1-(1-p)^n$; this presentation enables system-facing interpretations, e.g., bounding adversarial capability by operational query limits or compute budgets ($n\le N$), and selecting deployment-relevant thresholds that translate directly into risk under realistic sampling regimes.

\paragraph{Caveat on reproducibility.}
Complete end-to-end reproduction is sensitive to preprocessing details. \cite{hayes-etal-2025-measuring} does not describe the full data filtering and sampling procedure; our choices above follow the closest public descriptions but may not match prior pipelines exactly.

\section{Related Work}
\label{sec:related}
ML-Doctor~\cite{liu2022ml} and Amulet~\cite{waheed2025amulet} are open-source libraries designed to address and unify various privacy and security threats against ML models, including model extraction, model inversion (data reconstruction), membership and attribute inference attacks.
Regarding privacy attacks, none of these libraries incorporate methodologies for estimating empirical epsilon lower bounds.

The two main open-source libraries with a focus on privacy auditing are \emph{ML Privacy Meter}~\cite{mlprivacymeter}, and \emph{DP-auditorium}~\cite{kong2024dp}.
ML-Doctor~\cite{liu2022ml} and Amulet~\cite{waheed2025amulet} focus
 Membership Inference Attacks (MIAs) on classical ML models~\cite{tao2025range,tong2025much,ye2022enhanced,nasr2018comprehensive,rmia} developed from the library's maintainers research lab as well as empirical epsilon lower bounds via~\cite{steinke2024privacy}. PrivacyGuard offers an implementation of (i) SOTA MIA attacks based on shadow models~\cite{rmia,lira} together with lightweight MIA~\cite{calibmia} attacks which do not require additional model training and (ii) SOTA methodologies for evaluating the memorization of Large Language Models (LLMs) via probabilistic extraction attacks~\cite{hayes-etal-2025-measuring}. Further, PrivacyGuard provides a tighter (when compared to~\cite{steinke2024privacy}) and efficient one-run empirical epsilon estimation based on f-dp auditing~\cite{mahloujifar2025auditing}.

The methodology employed in \emph{DP-auditorium}~\cite{kong2024dp} does not rely on privacy inference attacks (e.g., MIA) as done in Privacy Guard. Instead, empirical privacy lower bounds are established by first measuring the distance (divergence) between the audited (randomized) mechanism’s output distributions, and then finding neighboring datasets where a mechanism generates output distributions maximizing such distance. By searching for worst case pairs of these neighboring datasets, DP-Auditorium mostly aims to find privacy bugs in the implementation of differentially private (black-box) algorithms. While privacy lower bounds are tight even for small theoretical epsilons, the library still needs to run several (in the order of 10s) trials of the target mechanisms. This is still not practical for  testing of expensive ML mechanisms (e.g., DP-SGD~\cite{abadi2016deep}) on real-world datasets.

\paragraph{Extraction Attacks for Language Models.}
With the advance of powerful language models recent years, new forms of privacy and security threats emerged against these large generative models. One of the most popular attacks is known as extraction attack, whose goal is to extract training data information from language models. By leveraging the power of modern language models, these extraction attacks do not rely on optimizing gradient and model weights. \cite{carlini2021extracting} first propose a prompting based extraction with known prefix information and show that PII information could be extracted from a GPT-2 trained with data containing these information. \cite{nasr2023} further applied the same technique on larger, production level language models and achieves non trivial extraction of data. More recently, \cite{hayes-etal-2025-measuring} extended this attack by using the token probability instead of the token themselves of output as a metric to evaluate extractability. Despite the emergence of these attacks, we are not aware of open source library that systematically implements these threats for language models.

\section{Conclusion}
PrivacyGuard provides a practical, modular framework for empirical privacy assessment of ML models. By supporting state-of-the-art and advanced privacy metrics, it enables researchers and practitioners to rigorously evaluate privacy risks in both traditional and generative ML systems. Its extensible design and robust testing ensure reliability and adaptability to evolving privacy challenges. PrivacyGuard is open source, inviting the community to contribute to the library's development to advance privacy research and responsible AI development.

\bibliographystyle{alpha}
\bibliography{main}

\end{document}